\documentclass[10pt,twocolumn,letterpaper]{article}

\usepackage{wacv}
\usepackage{times}
\usepackage{epsfig}
\usepackage{graphicx}
\usepackage{amsmath}
\usepackage{amssymb}
\usepackage{amsfonts,bm}
\usepackage{adjustbox}

\def\dist{{\mathcal{D}}}
\def\emK{{K}}
\def\evalpha{{\alpha}}
\def\exp{{\mathbb{E}}}
\def\mI{{\bm{I}}}
\def\mK{{\bm{K}}}
\def\mX{{\bm{X}}}
\def\mZ{{\bm{Z}}}
\def\sR{{\mathbb{R}}}
\def\sZ{{\mathbb{Z}}}
\def\valpha{{\bm{\alpha}}}
\def\vc{{\bm{c}}}
\def\vtheta{{\bm{\theta}}}
\def\vx{{\bm{x}}}

\wacvfinalcopy %

\usepackage[breaklinks=true,bookmarks=false]{hyperref}

\begin{document}

\title{Meta Learning for Few-Shot One-class Classification}

\author{Gabriel Dahia\\
Department of Computer Science, Federal University of Bahia (UFBA) \\
{\tt\small gdahia@protonmail.com}
\and
Maur\'icio Pamplona Segundo\thanks{Work done while at the Federal University of Bahia (UFBA).} \\
Institute for Artificial Intelligence (AI+X), University of South Florida
}

\maketitle

\begin{abstract}
We propose a method that can perform one-class classification given only a
  small number of examples from the target class and none from the others. We
  formulate the learning of meaningful features for one-class classification as
  a meta-learning problem in which the meta-training stage repeatedly simulates
  one-class classification, using the classification loss of the chosen
  algorithm to learn a feature representation. To learn these representations,
  we require only multiclass data from similar tasks. We show how the Support
  Vector Data Description method can be used with our method, and also propose
  a simpler variant based on Prototypical Networks that obtains comparable
  performance, indicating that learning feature representations directly from
  data may be more important than which one-class algorithm we choose. We
  validate our approach by adapting few-shot classification datasets to the
  few-shot one-class classification scenario, obtaining similar results to the
  state-of-the-art of traditional one-class classification, and that improves
  upon that of one-class classification baselines employed in the few-shot
  setting. Our code is available at \url{https://github.com/gdahia/meta_occ}
\end{abstract}

\section{Introduction}
\label{sec:introduction}

One-class classification algorithms are the main approach to detecting
anomalies from normal data but traditional methods scale poorly both in
computational resources and sample efficiency with the data dimensions.
Attempting to overcome these problems, previous work proposed using deep neural
networks to learn feature representations for one-class classification. While
successful in addressing some of the problems, they introduced other
limitations. One problem with these methods is that some of them optimize a
metric that is related, but different than their true one-class classification
objective (\textit{e.g.}, input reconstruction~\cite{dcae}). Other methods
require imposing specific structure to the models, like using generative
adversarial networks (GANs)~\cite{gan, anogan}, or removing biases and
restricting the activation functions for the network model~\cite{deep-svdd}.
GANs are notoriously hard to optimize~\cite{wgan, gan-opt}, and removing biases
restrict which functions the models can learn~\cite{deep-svdd}. Furthermore,
these methods require thousands of samples from the target class, only to
obtain results that are comparable to that of the traditional
baselines~\cite{deep-svdd}.

We propose a method that overcomes these problems if we have access to data
from related tasks. By using recent insights from the meta-learning community
on how to learn to learn from related tasks~\cite{maml, protonet}, we show that
it is possible to learn feature representations suitable for one-class
classification by optimizing an estimator of its classification performance.
This not only allows us to optimize the one-class classification objective
without any restriction to the model besides differentiability but also
improves the data efficiency of the underlying algorithm. Our method obtains
similar performance to traditional methods while using 1,000 times fewer data
from the target class, defining a trade-off in the availability of data from
related tasks and data from the target class.

For some one-class classification tasks, there are related tasks, and so our
method's requirement is satisfied. For example, in fraud detection, we could
use normal activity from other users and create related tasks that consist of
identifying if the activity came from the user or not, while still employing
and optimizing one-class classification.

\begin{figure*}[t]
  \begin{center}
    \centerline{\includegraphics[width=\textwidth]{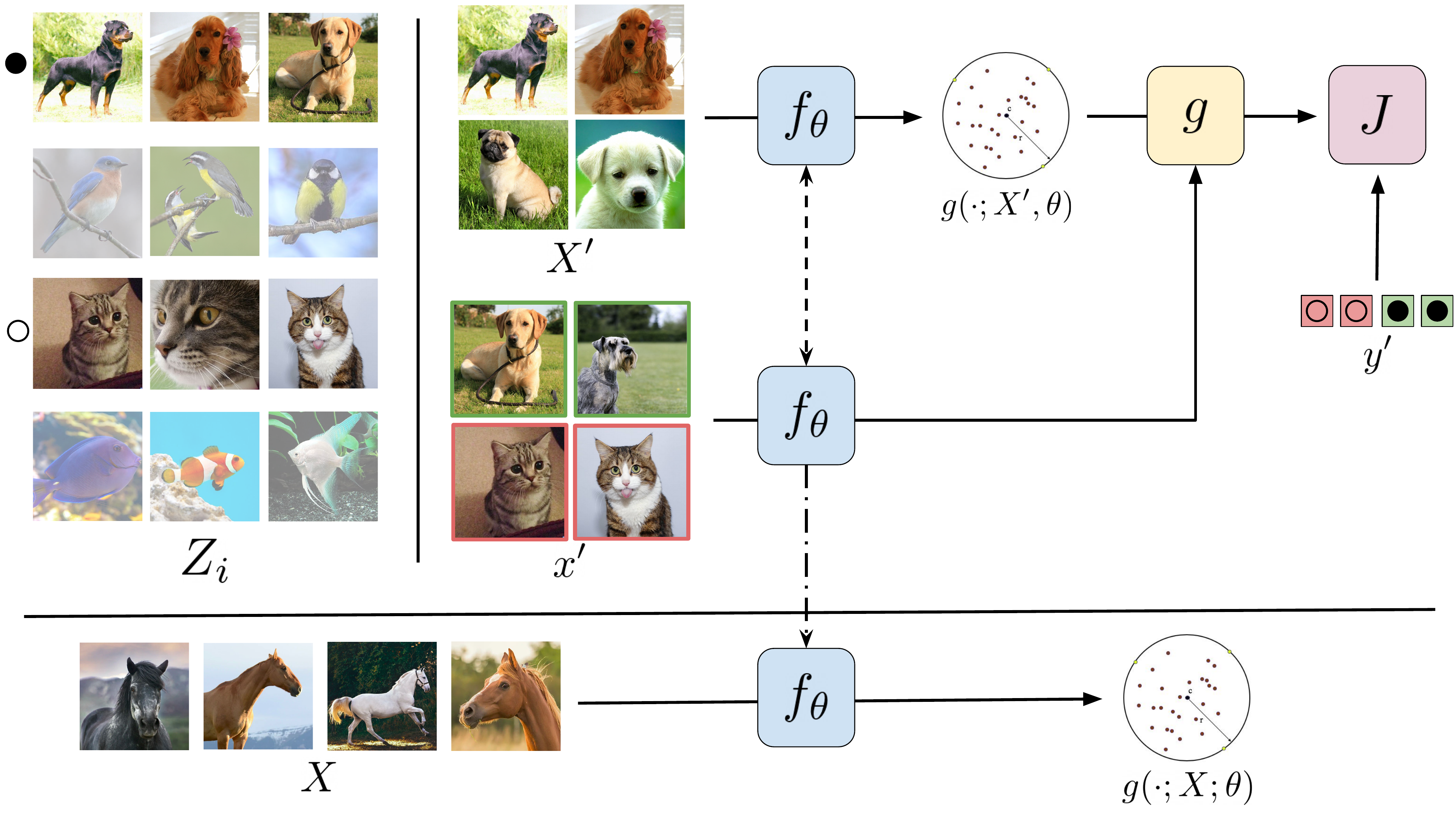}}
    \caption{Overview of the proposed method. During the meta-training stage,
    we emulate a training stage by first sampling $\mX'$ from a distribution
    that is similar to the one of our target class data $\mX$. In practice we
    use all examples from class $i$ - represented as the $\mZ_i$ sets in the
    figure - from a labeled dataset $\mZ$. We then sample a minibatch of pairs
    ${(\vx', y')}$ with $\vx'$ being an example and $y'$ a binary label
    indicating whether $\vx'$ belongs to the same class as the examples in
    $\mX'$, sampling again from sets $\mZ_i$. Then, we use a one-class
    classification algorithm (\textit{e.g.} SVDD) in the features resulting
    from applying $f_\vtheta$ on the examples of $\mX'$. We use the resulting
    classifier to classify each example's features $f_\vtheta(\vx')$ as
    belonging or not to the same class as $\mX'$, and compute the binary loss
    $J$ with the true labels $y'$. We optimize $f_\vtheta$ by doing gradient
    descent in the value of $J$ over many such tasks. After $f_\vtheta$ is
    learned, we run the same one-class classification algorithm on the
    resulting features, represented by the dashed-dotted arrow from the
    meta-training to the deployment stage, for $\mX$ in the true training
    stage, yielding the final one-class classification
    method.}\label{fig:pipeline}
  \end{center}
\end{figure*}

We describe an instance of our method, the \emph{Meta Support Vector Data
Description}, obtained by using the Support Vector Data Description
(SVDD)~\cite{svdd} as the one-class classification algorithm. We also simplify
this method to obtain a one-class classification variant of Prototypical
Networks~\cite{protonet}, which we call \emph{One-class Prototypical Network}.
Despite its simplicity, this method obtains comparable performance to Meta
SVDD\@. Our contributions thus are:
\begin{itemize}
  \item We show how to learn a feature representation for one-class
    classification (Section~\ref{sec:meta-svdd}) by defining an estimator for
    the classification loss of such algorithms (Section~\ref{sec:estimator}).
    We also describe how to efficiently backpropagate through the objective
    when the chosen algorithm is the SVDD method, so we can parametrize the
    feature representation with deep neural networks
    (Section~\ref{sec:grad-opt}). The efficiency requirement to train our model
    serves to make it work in the few-shot setting.
  \item We simplify Meta SVDD by replacing how the center of its hypersphere is
    computed. Instead of solving a quadratic optimization problem to find the
    weight of each example in the center's averaging, we remove the weighting
    and make the center the result of an unweighted average
    (Section~\ref{sec:protonet}). The resulting One-class Prototypical Networks
    are simpler, have lower computational complexity and more stable training
    dynamics than Meta SVDD.
  \item After that, we detail how our method conceptually addresses the
    limitations of previous work (Section~\ref{sec:related-work}). We also show
    that our method has promising empirical performance by adapting two
    few-shot classification datasets to the one-class classification setting
    and obtaining comparable results with the state-of-the-art of the many-shot
    setting (Section~\ref{sec:exps}). Our results indicate that learning the
    feature representations may compensate for the simplicity of replacing SVDD
    with feature averaging and that our approach is a viable way to replace
    data from the target class with labeled data from related tasks.
\end{itemize}

\section{Meta SVDD}\label{sec:meta-svdd}

The Support Vector Data Description (SVDD) method~\cite{svdd} computes the
hypersphere of minimum volume that contains every point in the training set.
The idea is that only points inside the hypersphere belong to the target class,
so we minimize the sphere's volume to reduce the chance of including points
that do not belong in the target class.

Formally, the radius ${R(\mX, \vc; \vtheta)}$ of the hypersphere centered at
$\vc \in \sR^d$ covering the training set $\mX$ transformed by ${f_\vtheta :
\sR^D \rightarrow \sR^d}$ is
\begin{equation}\label{eq:radius}
  R(\mX, \vc; \vtheta) = \max_{\vx \in \mX} \Vert f_\vtheta(\vx) - \vc \Vert.
\end{equation}
The SVDD objective is to find the center $\vc^*$ that minimizes the radius of
such a hypersphere, \textit{i.e.}
\begin{equation}\label{eq:center}
  \vc^* = \operatorname*{\arg\min}_{\vc} R(\mX, \vc; \vtheta).
\end{equation}
Finally, the algorithm determines that a point $\vx'$ belongs to the target
class if
\begin{equation}\label{eq:decision-criterion}
  \Vert f_\vtheta(\vx') - \vc^* \Vert \le R(\mX, \vc^*; \vtheta).
\end{equation}

The SVDD objective, however, does not specify how to optimize the feature
representation $f_\vtheta$. Previous approaches include using dimensionality
reduction with Principal Component Analysis (PCA)~\cite{deep-svdd}, using a
Gaussian kernel with the kernel trick~\cite{svdd}, or using features learned
with unsupervised learning methods, like deep belief networks~\cite{dbn-svm}.
We take a different approach: Our goal is to learn $f_\vtheta$ for the task,
and we detail how next.

\subsection{Meta-learning One-class Classification}\label{sec:estimator}
Our objective is to learn an $f_\vtheta$ such that the minimum volume
hypersphere computed by the SVDD covers only the samples from the target class.
We, therefore, divide the learning problem into two stages. In the
\emph{meta-training} stage, we learn the feature representation $f_\vtheta$.
Once we learn $f_\vtheta$, we use it to learn a one-class classifier using the
chosen algorithm (in this case, SVDD) from the data of the target class in the
\emph{training} stage. This is illustrated in Figure~\ref{fig:pipeline}.

Notice how both the decision on unseen inputs
(Equation~\ref{eq:decision-criterion}) and the hypersphere's center $\vc^*$
(Equation~\ref{eq:center}) depend on $f_\vtheta$. Perfectly learning
$f_\vtheta$ in the meta-training stage would map \emph{any} input distribution
into a space that can be correctly classified by SVDD, and would therefore not
depend on the given data $\mX$ nor on what is the target class; that would be
learned by the SVDD after transforming $\mX$ with $f_\vtheta$ in the subsequent
training stage. We do not know how to learn $f_\vtheta$ perfectly but the above
observation illustrates that we do not need to learn it with data from the
target class.

With that observation, we can use the framework of nested learning
loops~\cite{anil} to describe how we propose to learn $f_\vtheta$:
\begin{itemize}
  \item \emph{Inner loop}: Use $f_\vtheta$ to transform the inputs, and use
    SVDD to learn a one-class classification boundary for the resulting
    features.
  \item \emph{Outer loop}: Learn $f_\vtheta$ from the classification loss
    obtained with the SVDD.
\end{itemize}
We use the \emph{expected} classification loss in the outer loop. With this, we
can use data that comes from the same distribution as the data for the target
class, but with different classification tasks. To make this definition formal,
first, let $g$ be a one-class classification function parametrized by $\vtheta$
which receives as inputs a subset of examples from the target class $\mX'$ and
an example $\vx'$, and outputs the probability that $\vx'$ belongs to the
target class. For a suitable classification loss $J$, our learning loss is
\begin{equation}\label{eq:loss}
  \mathcal{L}(\vtheta) = \exp_{\mX' \sim \dist_\mX}[ \exp_{(\vx', y') \sim
  \dist_{\mZ|\mX'}}[J(g(\vx', \mX'; \vtheta), y')] ]
\end{equation}
where $y'$ is a binary label indicating whether $\vx'$ belongs to the same
distribution of $\mX'$ or not. The outer expectation of Equation~\ref{eq:loss}
defines a one-class classification task, and the inner expectation is over
labeled examples for this task (hence the dependency on $\mX'$ for the labeled
example distribution $\dist_{\mZ|\mX'}$). Since we do not have access to the
distribution $\dist_\mX$ nor we have access to $\dist_{\mZ | \mX}$, we
approximate it with related tasks. Intuitively, the closer the distribution of
the tasks we use to approximate it, the better our feature representation.

To compute this approximation in practice, we require access to a labeled
multiclass classification dataset ${\mZ = \{(\vx_1, y_1), \dots, (\vx_N,
y_N)\}}$, where $\vx_i \in \sR^D$ is the $i$\textsuperscript{th} element and
$y_i \in \sZ$ its label, that has a distribution similar to our dataset $\mX$,
but is disjoint from it (\textit{i.e.} none of the elements in $\mX$ are in
$\mZ$ and none of its elements belong to any of the classes in $\mZ$). Datasets
like $\mZ$ are common in the meta-learning or few-shot learning literature, and
their existence is a standard assumption in previous work~\cite{protonet, maml,
metaoptnet}. However, this restricts the tasks to which our method can be
applied to those that have such related data available.

We then create the datasets $\mZ_1, \dots, \mZ_k$ from $\mZ$ by separating its
elements by class, \textit{i.e.}
\begin{equation}\label{eq:dataset}
  \mZ_i = \{\vx_j \mid (\vx_j, i) \in \mZ\}.
\end{equation}
We create the required binary classification tasks by picking $\mZ_i$ as the
data for the target class, and the examples from $\mZ_j$, ${j \neq i}$, to be
the input data from the negative class. Finally, we approximate the
expectations in Equation~\ref{eq:loss} by first sampling mini-batches of these
binary classification tasks and then averaging over mini-batches of labeled
examples $\mZ'$ from each of the sampled tasks. By making each sampled $\mX'$
have few examples (\textit{e.g.} 5 or 20), we not only make our method scalable
but we also learn $f_\vtheta$ for \emph{few-shot one-class classification}.

In the next section, we define a model for $f_\vtheta$ and a way to optimize it
over Equation~\ref{eq:loss}.

\subsection{Gradient-based Optimization}\label{sec:grad-opt}
If we choose $f_\vtheta$ to be a neural network, it is possible to optimize it
to minimize the loss in Equation~\ref{eq:loss} with gradient descent as long as
$J$ and $g$ are differentiable and have meaningful gradients because of the
chain rule of calculus. $J$ can be the standard binary cross-entropy between
the data and model distributions~\cite{deep-learning-book}.

We also modify the SVDD to satisfy the requirements of the $g$ function.
Neither how it computes the hypersphere's center, by solving an optimization
problem (Equation~\ref{eq:center}), nor its hard, binary decisions
(Equation~\ref{eq:decision-criterion}) are immediately suitable for
gradient-based optimization.

To solve the hard, binary decisions problem, we adopt the approach of
Prototypical Networks~\cite{protonet} and consider the squared distance from
the features $f_\vtheta(\vx')$ to the center $\vc^*$ (the left-hand side of
Equation~\ref{eq:decision-criterion}) as the input logits for a logistic
regression model. Doing this not only solves the problem of uninformative
gradients coming from the binary outcomes of SVDD but also simplifies its
implementation in modern automatic differentiation/machine learning software,
\textit{e.g.} PyTorch~\cite{pytorch}. As our logits are non-negative, using the
sigmoid function $\sigma$ to convert logits into probabilities would result in
probabilities of at least 0.5 for every input, so we replace it with the
$\tanh$ and keep the binary cross-entropy objective otherwise unchanged.

As for how to compute $\vc^*$ in a differentiable manner, we can write it as
the weighted average of the input features
\begin{equation}\label{eq:center-dual}
  \vc^* = \sum_{i = 1}^n \evalpha_i f_\vtheta(\vx_i)
\end{equation}
where the weights $\valpha$ are the solution of the following quadratic
programming problem, which is the dual of the problem defined in
Equation~\ref{eq:center}~\cite{min-bounding-sphere-qp, svdd}
\begin{align}
  \max_\valpha~&~\valpha^T\mathrm{diag}(\mK) - \valpha^T \mK \valpha
  \label{eq:qp-max} \\
  \text{subject to}~&~\sum_{i = 1}^n \evalpha_i = 1 \label{eq:qp-eq} \\
                    &~0 \le \evalpha_i, i \in \{1, \dots, n\}
                    \label{eq:qp-ineq}
\end{align}
and
\begin{equation}\label{eq:kernel}
  \emK_{i, j} = f_\vtheta{(\vx_i)}^T f_\vtheta(\vx_j)
\end{equation}
is the kernel matrix of $f_\vtheta$ for input set $\mX$. Despite such quadratic
programs not having known analytical solutions and requiring a projection
operator to unroll its optimization procedure because of its inequality
constraints, the quadratic programming layer~\cite{optnet} can efficiently
backpropagate through its solution and supports GPU usage.

Still, the quadratic programming layer has complexity $O(m^3)$ for $m$
optimization variables~\cite{optnet}; in the case of Meta SVDD, $m$ is equal to
the number of examples in $\mX$ during training~\cite{metaoptnet}. As the size
of the network is constant, this is the overall complexity of performing a
training step in the model. Since we keep the number of examples small, 5 to
20, the runtime is dominated by the computation of $f_\vtheta$.

In practice, we follow previous work that uses quadratic programming
layers~\cite{metaoptnet} and we add a small stabilization value ${\lambda =
10^{-6}}$ to the diagonals of the kernel matrix (Equation~\ref{eq:kernel}),
\textit{i.e.}
\begin{equation}
  \mK' = \mK + \lambda \mI
\end{equation}
and we use $\mK'$ in Equation~\ref{eq:qp-max}. Not adding this stabilization
term results in failure to converge in some cases.

Using the program defined by objective~\ref{eq:qp-max}, and
constraints~\ref{eq:qp-eq} and~\ref{eq:qp-ineq} to solve SVDD also allows us to
use the kernel trick to make $\mK$ non-linear with regards to
$f_\vtheta$~\cite{svdd}. We believe this would not add much since using a deep
neural network to represent $f_\vtheta$ can handle the non-linearities that map
the input to the output, in theory.

SVDD~\cite{svdd} also introduce slack variables to account for outliers in the
input set $\mX$. Since our setting is few-shot one-class classification, we do
not believe these would benefit the method's performance because we think
outliers are unlikely in such small samples. We leave the analysis to confirm
or refute these conjectures to future work.

\section{One-class Prototypical Networks}\label{sec:protonet}

The only reason to solve the quadratic programming problem defined by
objective~\ref{eq:qp-max} and constraints~\ref{eq:qp-eq} and~\ref{eq:qp-ineq}
is to obtain the weights $\valpha$ for the features of each example in
Equation~\ref{eq:center-dual}.

We experiment with replacing the weights $\valpha$ in
Equation~\ref{eq:center-dual} by uniform weights $\evalpha_i = 1/n$. The center
$\vc^*$ then becomes a simple average of the input features
\begin{equation}
  \vc^* = \frac{1}{n} \sum_{i = 1}^n f_\vtheta(\vx_i)
\end{equation}
and we no longer require solving the quadratic program. The remainder of the
method, \textit{i.e.} its training objective, how tasks are sampled, etc,
remains the same. This avoids the cubic complexity in the forward pass, and the
destabilization issue altogether. We call this method One-class Prototypical
Networks because the method can be cast as learning binary Prototypical
Networks~\cite{protonet} with a binary cross-entropy objective.

Despite being a simpler method than Meta SVDD, we conjecture that learning
$f_\vtheta$ to be a good representation for One-class Prototypical Networks can
compensate its algorithmic simplicity so that performance does not degrade.

\section{Related work}\label{sec:related-work}

\subsection{One-class Classification}

The SVDD~\cite{svdd}, reviewed in Section~\ref{sec:meta-svdd}, is closely
related to the One-class Support Vector Machines (One-class SVMs)~\cite{ocsvm}.
Whereas the SVDD finds a hypersphere to enclose the input data, the One-class
SVM finds a maximum margin hyperplane that separates the inputs from the origin
of the coordinate system.  Like the SVDD, it can also be formulated as a
quadratic program, solved in kernelized form, and use slack variables to
account for outliers in the input data. In fact, when the chosen kernel is the
commonly used Gaussian kernel, both methods are equivalent~\cite{ocsvm}.

Besides their equivalence in that case, the One-class SVM more generally
suffers from the same limitations as the SVDD: it requires explicit feature
engineering (\textit{i.e.} it prescribes no way to formulate $f_\vtheta$), and
it scales poorly both with the number of samples and the dimension of the data.

In Section~\ref{sec:meta-svdd}, we propose to learn $f_\vtheta$ from related
tasks, which addresses the feature engineering problem. We also make it so that
it requires only a small set to learn the one-class classification boundary,
solving the scalability problem in the number of samples. Finally, by making
the feature dimension $d$ much smaller than $D$, we solve the scalability issue
regarding the feature dimensionality.

The limitations of SVDD and One-class SVMs led to the development of
\emph{deep} approaches to one-class classification, where the previous
approaches are known as \emph{shallow} because they do not rely on deep
(\textit{i.e.} multi-layered) neural networks for feature representation.

Most previous approaches that use deep neural networks to represent the input
feature for downstream use in one-class classification algorithms are trained
with a surrogate objective, like the representation learned for input
reconstruction with deep autoencoders~\cite{autoencoder}.

Autoencoder methods learn feature representations by requiring the network to
reconstruct inputs while preventing it to learn the identity function. These
are usually divided into an encoder, tasked with converting an input example
into an intermediate representation, and a decoder, that gets the
representation and must reconstruct the input~\cite{deep-learning-book}.

The idea is that if the identity function cannot be learned, then the
representation has captured semantic information of the input that is
sufficient for its partial reconstruction and other tasks. How the identity
function is prevented determines the type of autoencoder and many options
exist: by reducing the dimensions of or imposing specific distributions to the
intermediate representations, by adding a regularization term to the model's
objective, or by corrupting the input with noise~\cite{deep-learning-book}.

Philipp Seeb{\"{o}}ck \textit{et~al.}~\cite{dcae} train a deep convolutional
autoencoder (DCAE) in images for the target class, here healthy retinal image
data, and after that the decoder is ignored and a One-class SVM is trained on
the resulting intermediate representations. The main issue with this approach
is that the objective of autoencoder training does not assure that the learned
representations are useful for classification.

A related approach is to reuse features from networks trained for multiclass
classification. Oza and Patel~\cite{oc-cnn} remove the softmax layer of a
Convolutional Neural Network (CNN)~\cite{cnn} trained in the ImageNet
dataset~\cite{imagenet} as its feature extractor. The authors then train the
fully-connected layers of the pre-trained network alongside a new fully
connected layer tasked with discriminating between features from the target
class and data sampled from a spherical Gaussian distribution; the
convolutional layers are not updated.

AnoGANs~\cite{anogan} are trained as Generative Adversarial Networks~\cite{gan}
to generate samples from the target class. After that, gradient descent is used
to find the sample in the noise distribution that best reconstructs the unseen
example to be classified, which is equivalent to approximately inverting the
generator using optimization. The classification score is the input
reconstruction error, which assumes pixel-level similarity determines
membership in the target class.

Like our method, Deep SVDD~\cite{deep-svdd} attempts to learn feature
representations for one-class classification from the data using gradient-based
optimization with a neural network model. It consists of directly reducing the
volume of a hypersphere containing the features, and in that it is a deep
version of the original SVDD\@.

Deep SVDD's algorithm relies on setting the centers every few iterations with
the mean of the features from a forward pass instead of computing the minimum
bounding sphere. Since their objective is to minimize the volume of the
hypersphere containing the features, the algorithm must avoid the pathological
solution of outputting a constant function. This requires imposing
architectural constraints on the network, the stronger of which is that the
network's layers can have no bias terms. The authors also initialize the
weights with those of an encoder from a trained autoencoder. Neural network
models in our method have no such restrictions and do not require a
pre-training stage.

One advantage of Deep SVDD over our work is that it does not require data from
tasks from a similar distribution: it is trained only on the target class data.
While this is an advantage, there is a downside to it. It is not clear for us,
reading the paper describing Deep SVDD, how to know for how long to train a
Deep SVDD model, how to tune its many hyperparameters, or what performance to
expect of the method in unseen data. These are usually done with computing
useful metrics in a validation set. However, for Deep SVDD, the optimal value
can be reached for pathological solutions, so a validation set is not useful.

Ruff \textit{et~al.}~\cite{deep-svdd} prove that using certain activation
functions or keeping bias terms allow the model to learn the constant function
but they do not prove the reciprocate, \textit{i.e.} they do not prove that
constant functions cannot be learned by the restricted models. The authors also
do not analyze which functions are no longer learnable when the model is
restricted as such. For Meta SVDD, on the other hand, the related tasks give
predictive measures of metrics of interest, allow tuning hyperparameters, and
early stopping.

\subsection{Few-shot Learning}

The main inspiration for the ideas in our paper besides Deep SVDD came from the
field of meta-learning, in particular, that of few-shot classification.
Prototypical Networks~\cite{protonet} are few-shot classifiers that create
prototypes from few labeled examples and use their squared distances to an
unseen example as the logits to classify it as one of their classes. We first
saw the idea of learning the feature representation from similarly distributed
tasks and of using the squared distances in this paper. They also propose
feature averaging as a way to summarize class examples and show its competitive
performance despite its simplicity; One-class Prototypical Networks are the
one-class variant of this method.

Recently, Lee \textit{et~al.}~\cite{metaoptnet} proposed to learn feature
representations for few-shot classification convex learners, including
multi-class Support Vector Machines~\cite{svm}, with gradient-based
optimization. Their work is similar to ours in its formulation of learners as
quadratic programs, and in solving these with quadratic programming layers but
it does not address one-class classification.

\section{Experiments}\label{sec:exps}

\subsection{Evaluation Protocol}\label{sec:protocol}

Our first experiment is an adaptation of the evaluation protocol of Deep
SVDD~\cite{deep-svdd} to the few-shot setting to compare Meta SVDD with
previous work. The original evaluation protocol consists of picking one of the
classes of the dataset, training the method in the examples in the training set
(using the train-test split proposed by the maintainers), and using all the
examples in the test set to compute the mean and standard deviation of the Area
under the curve (AUC) of the trained classifier over 10 repetitions in the
MNIST~\cite{mnist} and CIFAR-10~\cite{cifar10} datasets.

We modified the protocol because there are only 10 classes in these datasets,
which is not enough for meta-learning one-class classifiers. This illustrates
the trade-off introduced by our approach: Despite requiring many fewer examples
per class, it requires many more classes. Our modifications are only to address
the number of classes and we tried to keep the protocol as similar as possible
to make the results more comparable.

The first modification is the replacement of CIFAR-10 by the CIFAR-FS
dataset~\cite{cifar-fs}, a new split of CIFAR-100 for few-shot classification
in which there is no class overlap between the training, validation and test
sets. CIFAR-FS has 64 classes for training, 16 for validating, and 20 for
testing, and each class has 600 images.

No such split is possible for MNIST because there is no fine-grained
classification like in the case of the CIFAR-10 and CIFAR-100 datasets.
Therefore, we use the Omniglot dataset~\cite{omniglot}, which is considered the
``transposed'' version of the MNIST dataset because it has many classes with
few examples instead of the many examples in the 10 classes of MNIST\@. This
dataset consists of 20 images of each of its 1623 handwritten characters, which
are usually augmented with four multiples of $90^\circ$ to obtain ${1623 \times
4 = 6492}$ classes~\cite{matching-net, cifar-fs, protonet, maml}. We follow the
pre-processing and dataset split proposed by
Vinylas~\textit{et~al.}~\cite{matching-net} by resizing the images to ${28
\times 28}$ pixels, and using 4800 classes for training and 1692 for testing,
which is nowadays standard in few-shot classification work~\cite{maml,
protonet, cifar-fs}.

Another modification is that since there are only 10 classes in MNIST and
CIFAR-10, Deep SVDD~\cite{deep-svdd} reports the AUC metrics for each class.
This is feasible for CIFAR-FS, which has 20 testing classes, but not for
Omniglot, which has 1692. We summarize these statistics by presenting the
minimum, median, and maximum mean AUC alongside their standard deviations.

The last modification is in the number of elements per class in the test set
evaluation. Since there are many classes and we are dealing with few-shot
classification, we use only two times the number of examples in $\mX$ for the
target and for the negative class, \textit{e.g.} if the task is 5-shot
learning, then there are 10 examples from the target class and 10 examples from
the negative class for evaluation.

To better compare the previous methods with ours in the few-shot setting, we
evaluate the state-of-the-art method for general deep one-class classification,
Deep SVDD~\cite{deep-svdd}, in our modified protocol. We run the evaluation
protocol in CIFAR-FS using only 5 images for training, and we evaluate it using
10 images from the target class and 10 images from a negative class, and we do
this 10 times for each pair of the 20 test classes to compute mean and standard
deviation statistics for the AUC\@. We don't do this for Omniglot because it
would require training more than 1692 Deep SVDD models.

We also conduct a second experiment, based on the standard few-shot
classification experiment in which we evaluate the mean 5-shot one-class
classification accuracy over 10,000 episodes of tasks consisting of 10 examples
from the target class and 10 examples from the negative class. We use this
experiment to compare with a shallow baseline, PCA and Gaussian kernel
One-class SVM~\cite{ocsvm}, and One-class Prototypical Network. We use the
increased number of episodes to compute 95\% confidence intervals like previous
work for few-shot multiclass classification~\cite{cifar-fs, metaoptnet}.

\begin{table}[t]
  \begin{center}
    \begin{adjustbox}{width=0.48\textwidth}
      \begin{tabular}{|c||l|c|r||l|c|c|r|}
        \hline
        &
        Dataset
        &
        DCAE &
        \begin{tabular}{c}
          Deep \\ SVDD
        \end{tabular} &
        Dataset &
        \begin{tabular}{c}
          Deep \\ SVDD
        \end{tabular} &
        \begin{tabular}{c}
          One-Class \\ Protonet
        \end{tabular} &
        \begin{tabular}{c}
          Meta \\ SVDD
        \end{tabular} \\
        \hline\hline
        Min. &
        &
        78.2 $\pm$ 2.7 &
        {\bf 88.5 $\pm$ 0.9} &
        &
        &
        {\bf 89.0 $\pm$ 0.2} &
        {\bf 88.6 $\pm$ 0.4} \\
        Med. &
        MNIST &
        86.7 $\pm$ 0.9 &
        94.6 $\pm$ 0.9 &
        Omniglot &
        -- &
        {\bf 99.5 $\pm$ 0.0} &
        {\bf 99.5 $\pm$ 0.0} \\
        Max. &
        &
        98.3 $\pm$ 0.6 &
        99.7 $\pm$ 0.1 &
        &
        &
        {\bf 100.0 $\pm$ 0.0} &
        {\bf 100.0 $\pm$ 0.0} \\
        \hline
        Min. &
        &
        51.2 $\pm$ 5.2 &
        50.8 $\pm$ 0.8 &
        &
        47.9 $\pm$ 4.9 &
        {\bf 60.2 $\pm$ 3.4} &
        {\bf 59.0 $\pm$ 5.7} \\
        Med. &
        CIFAR-10 &
        58.6 $\pm$ 2.9 &
        65.7 $\pm$ 2.5 &
        CIFAR-FS &
        64.0 $\pm$ 5.0 &
        {\bf 72.7 $\pm$ 3.0} &
        {\bf 71.0 $\pm$ 4.0} \\
        Max. &
        &
        76.8 $\pm$ 1.4 &
        75.9 $\pm$ 1.2 &
        &
        {\bf 92.4 $\pm$ 2.3} &
        {\bf 90.1 $\pm$ 2.3} &
        {\bf 92.5 $\pm$ 1.7} \\
        \hline
      \end{tabular}
    \end{adjustbox}
  \end{center}
  \caption{Minimum, median and maximum mean AUC alongside their standard
  deviation for one-class classification methods for 10 repetitions. We
  highlight in boldface the highest mean and others which are within one
  standard deviation from it. The results for the many-shot baselines in MNIST
  and CIFAR-10 are compiled from the table by
  Ruff~\textit{et~al.}~\cite{deep-svdd}. The results for Omniglot and CIFAR-FS
  are for 5-shot one-class classification.}\label{tab:auc}
\end{table}

\subsection{Setup}

We parametrize $f_\vtheta$ with the neural network architecture model
introduced by Vinyals~\textit{et~al.}~\cite{matching-net} that is commonly used
in other few-shot learning work~\cite{maml, protonet}. There are four
convolutional blocks with number of filters equal to 64, and each block is
composed of a ${3 \times 3}$ kernel, stride 1, ``same'' 2D convolution, batch
normalization~\cite{batchnorm}, followed by ${2 \times 2}$ max-pooling and ReLU
activations~\cite{relu}.

We implemented the neural network using PyTorch~\cite{pytorch} and the
$\texttt{qpth}$ package~\cite{optnet} for the quadratic programming layer. We
also used Scikit-Learn~\cite{sklearn} and NumPy~\cite{numpy} to compute
metrics, implement the shallow baselines and for miscelaneous tasks, and
Torchmeta~\cite{torchmeta} to sample mini-batches of tasks, like described in
Section~\ref{sec:estimator}.

We optimize both Meta SVDD and One-class Prototypical Networks using stochastic
gradient descent~\cite{sgd} on the objective defined in
Section~\ref{sec:estimator} and Equation~\ref{eq:loss} with the Adam
optimizer~\cite{adam}. We use a constant learning rate of ${5 \times 10^{-4}}$
over mini-batches of tasks of size 16, each having set $\mX'$ with 5 examples,
and set $\mZ'$ with 10 examples from the target class and 10 examples from a
randomly picked negative class. The learning rate value was the first one we
tried, so no tuning was required. We picked the task batch size that performed
better in the validation set when training halts; we tried sizes ${\{2, 4,
\dots, 32\}}$. We evaluate the performance in the validation set with 95\%
confidence intervals of the model's accuracy in 500 tasks randomly sampled from
the validation sets, and we consider that a model is better than another if the
lower bound of its confidence interval is greater, or if its mean is higher
when the lower bounds are equal up to 5 decimal points. Early stopping halts
training when performance in the validation set does not increase for 10
evaluations in a row, and we use the model with higher performance in the
validation set. We evaluate the model in the validation set every 100 training
steps.

The results for the few-shot experiment with Deep SVDD are obtained modifying
the code made available by the
authors\footnote{\url{https://github.com/lukasruff/Deep-SVDD-Pytorch}}, keeping
the same hyperparameters.

For the few-shot baseline accuracy experiment with PCA and One-class SVMs with
Gaussian kernel, we use the grid search space used by the experiments in prior
work~\cite{deep-svdd}: $\gamma$ is selected from ${\{2^{-10}, 2^{-9}, \dots,
2^{-1}\}}$, and $\nu$ is selected from ${\{0.01, 0.1\}}$. Furthermore, we give
the shallow baseline an advantage by evaluating every parameter combination in
the test set and reporting the best result.

\begin{figure}[t]
  \begin{center}
    \centerline{\includegraphics[width=0.45\textwidth]{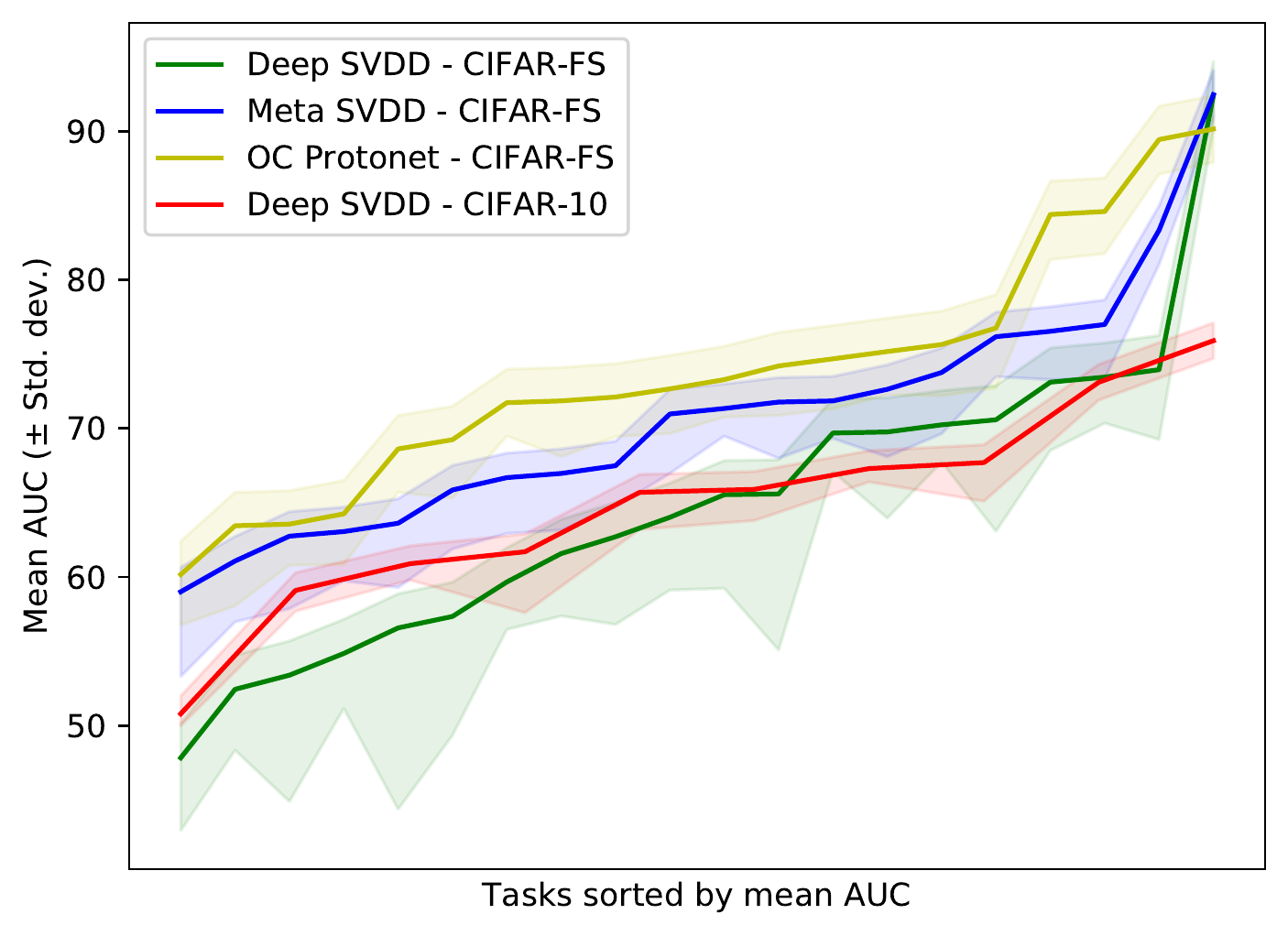}}
  \end{center}
  \vspace{-20pt}
  \caption{Mean AUC with shaded standard deviations for tasks in CIFAR datasets
  sorted by increasing mean value. Comparing Deep SVDD across datasets and
  protocols shows that the modified protocol is reasonable to evaluate few-shot
  one-class classification because the trend in task difficulty is similar.
  Within the few-shot protocol in CIFAR-FS, meta one-class classification are
  numerically superior, show less variance and can be meta-trained once for all
  tasks, with simple adaptation for unseen tasks, but require related task
  data.}\label{fig:cifar_plot}
\end{figure}

\subsection{Results}\label{sec:results}

We reproduce the results reported for Deep SVDD~\cite{deep-svdd} and its
baselines alongside the results for 5-shot Meta SVDD and One-class Prototypical
Networks, and our experiment with 5-shot Deep SVDD in Table~\ref{tab:auc}.
Figure~\ref{fig:cifar_plot} also provides mean AUC with shaded standard
deviations for the results in the CIFAR dataset variants.

While the results from different datasets are not comparable due to the
differences in setting and application listed in Section~\ref{sec:protocol},
they show that the approach has similar performance to the many-shot
state-of-the-art in terms of AUC\@. Figure~\ref{fig:cifar_plot} shows that when
we sort the mean AUCs for CIFAR-10 and CIFAR-FS, the performance from hardest
to easier tasks exhibit similar trends despite these differences, and that the
modifications to the protocol are reasonable.

This experiment is evidence that our method is able to reduce the required
amount of data from the target class in case we have labeled data from related
tasks. Note that it is not the objective of our experiments to show that our
method has better performance than previous approaches, since they operate in
different settings, \textit{i.e.} few-shot with related tasks and many-shot
without them.

The comparison with Deep SVDD in the few-shot scenario gives further evidence
of the relevance of our method: both Meta SVDD and One-Class Prototypical
Networks obtain higher minimum, and median AUC than Deep SVDD\@. Another
advantage is that we train $f_\vtheta$ once in the training set of Omniglot or
CIFAR-FS, and learn only either the SVDD or the average on each of the sets
$\mX$ in the test set. We also obtain these results without any pre-training,
and we have established a clear validation procedure to guide hyperparameter
tuning and early stopping.

These results also show we can train a neural network for $f_\vtheta$ without
architectural restrictions to optimize a one-class classification objective
whereas other methods either require feature engineering, optimize another
metric, or impose restrictions on the model architecture to prevent learning
trivial functions.

\begin{table}[t]
  \begin{center}
    \begin{adjustbox}{width=\columnwidth}
      \begin{tabular}{|l|c|c|r|}
        \hline Dataset
        &
        PCA+SVM &
        \begin{tabular}{c}
          One-class \\ Protonet
        \end{tabular} &
        Meta SVDD \\
        \hline\hline
        Omniglot &
        50.64 $\pm$ 0.10\% &
        {\bf 94.68 $\pm$ 0.17}\% &
        {\bf 94.33 $\pm$ 0.19}\% \\
        \hline
        CIFAR-FS &
        54.77 $\pm$ 0.31\% &
        {\bf 67.67 $\pm$ 0.39}\% &
        64.95 $\pm$ 0.37\% \\
        \hline
      \end{tabular}
    \end{adjustbox}
  \end{center}
  \caption{Mean accuracy alongside 95\% confidence intervals computed over
  10,000 tasks for Gaussian kernel One-class SVM with PCA, Meta SVDD and
  One-class Protoypical Networks. The results with highest mean and those with
  overlapping confidence interval with it are in boldface. We report the best
  result for the One-class SVM in its parameter search space, which gives it an
  advantage over the other two methods. Despite employing a simpler algorithm
  for one-class classification, One-class Prototypical networks obtain
  equivalent accuracy for Omniglot and better accuracy for CIFAR-FS than Meta
  SVDD\@. This indicates that learning feature representations is more
  important than which one-class classification algorithm we
  use.}\label{tab:acc}
\end{table}

The results for our second experiment, comparing the accuracies of Meta SVDD, a
shallow baseline and One-class Prototypical Networks are presented in
Table~\ref{tab:acc}.

In this experiment, we can see an increase from almost random performance to
almost perfect performance for both methods when compared to the shallow
baseline in Omniglot. Both methods for few-shot one-class classification that
use related tasks have equivalent performance in Omniglot. The gain is not as
significant for CIFAR-FS but more than 10\% in absolute for both methods, which
shows they are a marked improvement over the shallow baseline.

Comparing the two proposed methods, we observe the unexpected result that the
simpler method, One-class Prototypical Networks, has equivalent accuracy in the
Omniglot experiment, and \emph{better} accuracy in the CIFAR-FS experiment.
This indicates that learning the feature representation directly from data
might be more important than the one-class classification algorithm we choose,
and the increased complexity of using SVDD over simple averaging does not
translate into improved performance in this setting.

We have also attempted to run this same experiment in the \textit{mini}ImageNet
dataset~\cite{matching-net}, a dataset for few-shot learning using the images
from the ImageNet dataset~\cite{imagenet}. The accuracy in the validation set,
however, never rose above 50\%. One of the motivations of introducing CIFAR-FS
was that there was a gap in the challenge between training models in Omniglot
and \textit{mini}ImageNet and that successfully training models in the latter
took hours~\cite{cifar-fs}. Since none of the previous methods attempted
solving ImageNet level datasets, and the worst performance in datasets from
CIFAR is already near random guessing, we leave the problem of training
one-class classification algorithms in this dataset open for future work.

Finally, we have run a small variation of the second experiment in which the
number of examples in $\mX$ is greater than during training, using 10 examples
instead of 5. The results stayed within the accuracy confidence intervals for
5-shot for both models in this 10-shot deployment scenario.

\section{Conclusion}

We have described a way to learn feature representations so one-class
classification algorithms can learn decision boundaries that contain the target
class from data, optimizing an estimator of its true objective. Furthermore,
this method works with 5 samples from the target class with performance similar
to the state-of-the-art in the setting where target class data is abundant, and
better when the many-shot state-of-the-art method is employed in the few-shot
setting. We also provide an experiment that shows that using a simpler
one-class classification yields comparable performance, displaying the
advantages of learning feature representations directly from data.

One possibility to replace the main requirement of our method with a less
limiting one would be the capability of generating related tasks from unlabeled
data. A simple approach in this direction could be using weaker learners to
define pseudolabels for the data. Doing this successfully would increase the
number of settings where our method can be used significantly.

The main limitations of our method besides the requirement of the related tasks
are the destabilization of the quadratic programming layer, which we solved by
adding a stabilization term to the diagonal of the kernel matrix or by
simplifying the one-class classification algorithm to use the mean of the
features, and its failure to obtain meaningful results in the
\textit{mini}ImageNet dataset.

We believe not only finding solutions to these limitations should be
investigated in future work but also other questions left open in our work,
like confirming our hypothesis that introducing slacks would not benefit Meta
SVDD.

Other directions for future work are extending our method for other settings
and using other one-class classification methods besides SVDD\@. Tax and
Duin~\cite{svdd} also detail a way to incorporate negative examples in the SVDD
objective, so we could try learning $f_\vtheta$ using this method and to
minimize the hypersphere's volume instead of converting SVDD into a binary
classification problem that uses the unseen examples' distances to the center
as logits.

{\small
\bibliographystyle{ieee_fullname}
\bibliography{main}
}

\end{document}